\documentclass{article}

\usepackage[preprint]{neurips_2023}

\setcitestyle{numbers,square,comma}

\usepackage[utf8]{inputenc} % allow utf-8 input
\usepackage[T1]{fontenc}    % use 8-bit T1 fonts
\usepackage{hyperref}       % hyperlinks
\usepackage{url}            % simple URL typesetting
\usepackage{booktabs}       % professional-quality tables
\usepackage{amsfonts}       % blackboard math symbols
\usepackage{nicefrac}       % compact symbols for 1/2, etc.
\usepackage{microtype}      % microtypography
\usepackage{xcolor}         % colors
\usepackage{caption}
\usepackage{subcaption}
\usepackage{amsmath}
\usepackage{graphicx}
\usepackage{comment}
\usepackage{icomma}

\definecolor{citecolor}{RGB}{0,0,128} % Dark blue 

\hypersetup{
    colorlinks=true,
    linkcolor={citecolor},
    citecolor={citecolor}
}
\usepackage{cleveref}

\usepackage{pifont}
\usepackage{newunicodechar}
\newunicodechar{✓}{\ding{51}}
\newunicodechar{✗}{\ding{55}}

\definecolor{electricpurple}{rgb}{0.75, 0.0, 1.0}

\title{CLearViD: Curriculum Learning for Video Description}

\author{%
  Cheng-Yu Chuang \\
  Department of Computer Science \\
  San Francisco State University \\
  \texttt{cchuang2@sfsu.edu}
  \And
  Pooyan Fazli \\
  School of Arts, Media, and Engineering\\
  Arizona State University \\
  \texttt{pooyan@asu.edu} 
}

\begin{document}

\maketitle

\begin{abstract}

Video description entails automatically generating coherent natural language sentences that narrate the content of a given video.\ We introduce CLearViD, a transformer-based model for video description generation that leverages curriculum learning to accomplish this task. In particular, we investigate two curriculum strategies: (1) progressively exposing the model to more challenging samples by gradually applying a Gaussian noise to the video data, and (2) gradually reducing the capacity of the network through dropout during the training process. These methods enable the model to learn more robust and generalizable features. Moreover, CLearViD leverages the Mish activation function, which provides non-linearity and non-monotonicity and helps alleviate the issue of vanishing gradients. Our extensive experiments and ablation studies demonstrate the effectiveness of the proposed model. The results on two datasets, namely ActivityNet Captions and YouCook2, show that CLearViD significantly outperforms existing state-of-the-art models in terms of both accuracy and diversity metrics. 

\end{abstract}

\section{Introduction} 
\label{Introduction}

Video description is the task of automatically generating coherent natural language sentences that narrate the content of a given video, such as actions, characters, scene changes, interactions, etc. It has applications in video summarization, content retrieval, human-robot interaction, scene understanding, and accessibility for blind and low vision individuals~\cite{chi2020yuksel,yuksel2020human,bodi_2021}. Despite the high interest in this task and the emergence of new approaches and datasets~\cite{gella2018dataset, krishna2017dense, zhou2018towards}, video description remains a highly challenging problem. 

Curriculum learning involves presenting training examples to the model in a specific order or curriculum, gradually increasing the difficulty of the examples as training progresses~\cite{bengio2009curriculum}. The concept of curriculum learning is inspired by the way humans learn, starting with simpler concepts before moving on to more complex ones. By initially exposing the model to easier examples, it can build a solid foundation and gradually increase its learning capacity. As training progresses, more difficult examples are introduced, allowing the model to refine its understanding and handle complex patterns or concepts.

In traditional machine learning, training examples are typically randomly or sequentially presented to the model during training. However, curriculum learning proposes a different approach by carefully designing the order in which the examples or tasks are presented. The curriculum defines a sequence of tasks or examples that the model needs to learn, starting from easier instances and progressing to more challenging ones. Curriculum learning has been applied to various machine learning tasks, including image classification~\cite{chen2015webly}, object detection~\cite{10.1007/s11263-018-1112-4}, natural language processing~\cite{platanios2019competence}, and reinforcement learning~\cite{narvekar2020curriculum}. It has been shown to improve learning efficiency, convergence speed, and generalization performance~\cite{hacohen2019power}. However, designing an effective curriculum can be a challenging task and may require domain knowledge, manual selection, or heuristic rules. In this work, we introduce CLearViD (\textbf{C}urriculum \textbf{Lear}ning for \textbf{Vi}deo \textbf{D}escription), a transformer-based model for automatic video description generation. CLearViD leverages a mixed curriculum approach consisting of two strategies: (1) curriculum learning by noise and (2) curriculum learning by dropout.

\begin{figure}[t]
\centering
\begin{subfigure}{0.32\textwidth}
     \includegraphics[width=\textwidth]{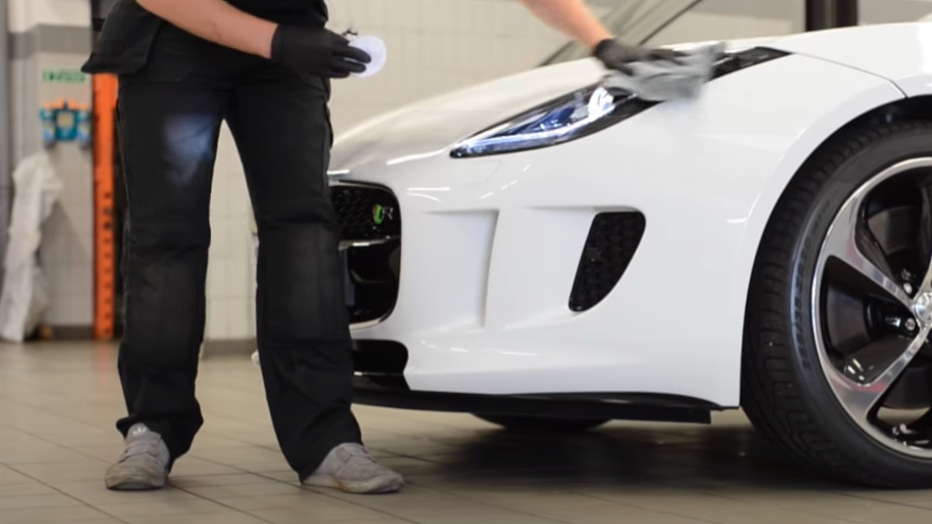}
     \caption{Easy}
     \label{fig:noise-easy}
\end{subfigure}
\begin{subfigure}{0.32\textwidth}
    \includegraphics[width=\textwidth]{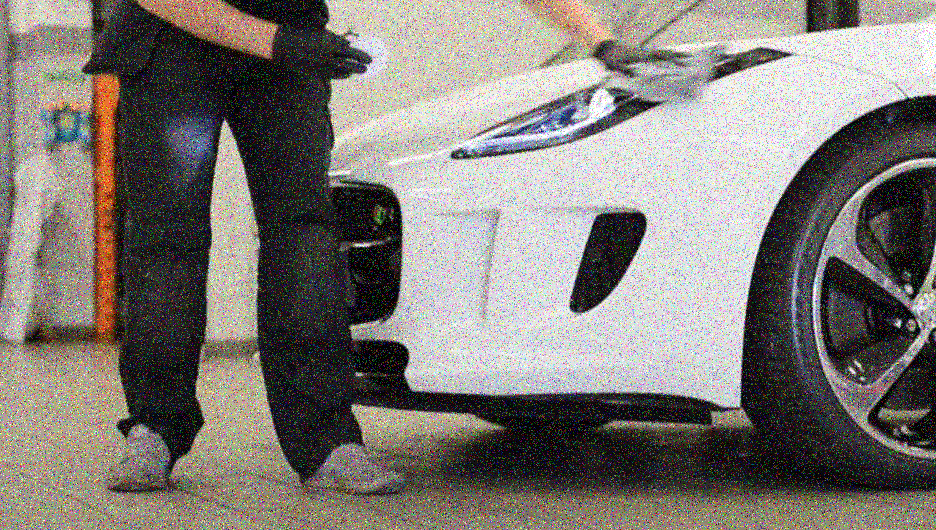}
    \caption{Medium}
    \label{fig:noise-medium}
\end{subfigure}
\begin{subfigure}{0.32\textwidth}
    \includegraphics[width=\textwidth]{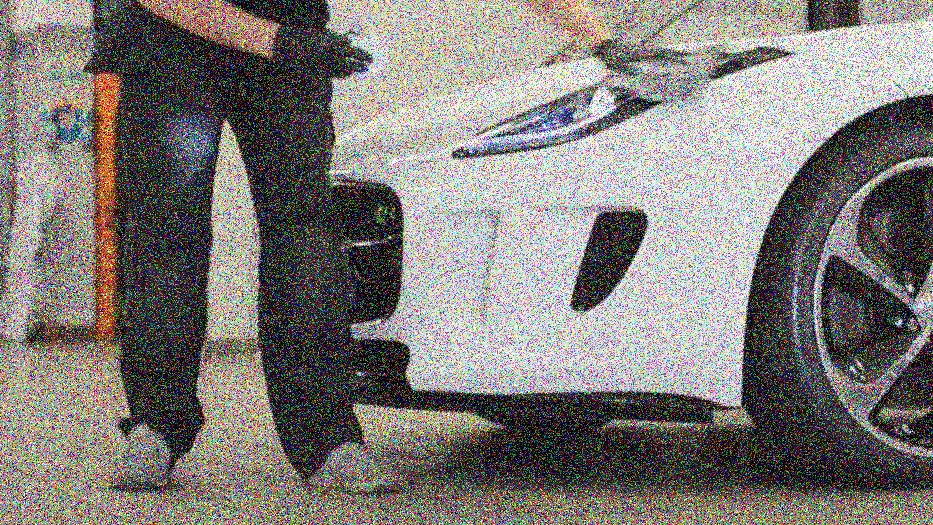}
    \caption{Hard}
    \label{fig:noise-hard}
\end{subfigure}
    \caption{Curriculum learning by noise. During the training process, we gradually increase the amount of Gaussian noise applied to the training set.}
    \label{fig:noise-demo}
\end{figure}

In curriculum learning by noise, the added noise can take different forms depending on the specific problem and the nature of the data. One common approach is to apply Gaussian noise, where random values from a Gaussian distribution are added to the training data. This introduces random perturbations to the input, making the model more robust to variations and noise during the test. The noise is gradually increased over the course of training, aligning with the curriculum learning principle of starting with easier examples and gradually exposing the model to more challenging ones (Figure~\ref{fig:noise-demo}). The introduction of noise forces the model to learn in a more robust and adaptive manner, improving its generalization capabilities and overall performance. Videos captured in real-life scenarios often contain various sources of noise, such as camera jitter, motion blur, and varying lighting conditions. By incorporating noise into the training data, curriculum learning can expose the model to a more realistic range of video quality and visual variations, enabling it to become robust to the challenges posed by noisy videos. 

Dropout is a regularization technique that randomly drops out a certain portion of the neurons during training to prevent overfitting and encourage the model to learn more robust features. In curriculum learning by dropout~\cite{morerio2017curriculum}, the dropout rate is dynamically adjusted during training to gradually increase the difficulty of the learning task. The idea is to start with a low dropout rate, allowing the model to initially use all of its neurons and learn the task in an easier setting. As training progresses, the dropout rate is increased, forcing the model to rely on a smaller subset of neurons and learn with a more limited capacity. This encourages the model to generalize beyond memorization, as it must now learn to recognize patterns and features with limited resources.

Another strategy for improving video description models is the careful selection of activation functions. The Mish activation function~\cite{misra2019mish} is characterized by its non-monotonic behavior, providing a smooth and non-linear transformation of the input data. These features allow Mish to capture intricate patterns and fine-grained details present in video content. In video description models, where the goal is to generate natural language descriptions for videos, the ability to recognize and capture such patterns can significantly enhance the model's understanding of the visual and temporal cues within the videos. This, in turn, can lead to more accurate and descriptive captions. Furthermore, the Mish activation function is continuously differentiable, making it suitable for backpropagation and gradient-based optimization algorithms. The smoothness and differentiability of Mish facilitate stable and efficient training of video description models. This helps in mitigating issues like vanishing gradients and accelerates the convergence of the model during training.

CLearViD, our proposed video description model, incorporates curriculum learning by noise, curriculum learning by dropout, and the Mish activation function to enhance its performance. We conducted extensive experiments and ablation studies on two benchmark datasets, namely ActivityNet Captions~\cite{krishna2017dense} and YouCook2~\cite{zhou2018towards}, to evaluate the effectiveness of CLearViD. The results of our experiments demonstrate that CLearViD significantly outperforms SOTA models across various evaluation metrics, indicating its superiority in generating high-quality and diverse video descriptions.

In summary, the main contributions of the paper are as follows:
\begin{enumerate}

\item designing a mixed curriculum by noise and dropout strategy to enable the model to learn more robust and generalizable features in video content,

\item investigating the effectiveness of the Mish activation function in enhancing the performance of the video description model, and

\item conducting comprehensive experiments and ablation studies on two widely-used datasets to validate the performance of the proposed model using standard evaluation metrics, focusing on both the accuracy and diversity of the generated video descriptions.

\end{enumerate}

\section{Related work}

\subsection{Video Description}

Early video description models mainly relied on handcrafted features and rule-based systems to generate descriptions~\cite{barbu2012video, kojima2002natural}. These methods often involved pre-segmenting videos into scenes or keyframes and using techniques such as template matching~\cite{guadarrama2013youtube2text}, object recognition~\cite{das2013thousand}, and syntactic parsing~\cite{srihari1995automatic} to generate descriptions. While these approaches provided initial insights into video description, they lacked the ability to generate diverse and contextually rich descriptions. The advent of sequence-to-sequence models, specifically based on encoder-decoder architectures~\cite{venugopalan-etal-2015-translating, yu2016video, yao2015describing, venugopalan2015sequence}, enabled significant advancements in the field of video description. Encoder-decoder architectures are generally divided into two stages: 1) visual content extraction or the encoding stage and 2) text generation or
the decoding stage. For encoding, convolutional neural networks (CNNs) are used to learn visual features, and for decoding, different variations of recurrent neural networks (RNNs), such as long short-term memory (LSTM) and gated recurrent unit (GRU) networks are used for language modeling and text generation. 

Inspired by the success of Transformer models in natural language processing tasks~\cite{vaswani2017attention}, researchers have investigated their application to video description~\cite{zhou2018end, lei2020mart, sun2019videobert}. Transformers leverage self-attention mechanisms to capture long-range dependencies and contextual information, making them effective in modeling the relationships between video frames or regions and generating coherent descriptions. These models have demonstrated SOTA performance and the ability to handle long videos and complex visual scenes. Recent advancements in video description have focused on incorporating multimodal information, such as visual features~\cite{yamazaki2023vltint}, audio features~\cite{iashin2020multi}, and semantic embeddings~\cite{dong2023semantic}, to enhance the quality and richness of generated descriptions. Multimodal fusion techniques, including early fusion~\cite{aafaq2022dense}, late fusion~\cite{xu2018dual, lei2020tvr}, and attention-based fusion~\cite{wu2018hierarchical, hori2017attention}, have been explored to effectively integrate different modalities and leverage their complementary information for improved video understanding and description generation.

\subsection{Curriculum Learning}

Bengio et al.~\cite{bengio2009curriculum} proposed curriculum learning to gradually increase training data's complexity to improve model generalization. 
Since then, curriculum learning has been successfully applied to various vision and language tasks, including image classification~\cite{chen2015webly}, object detection~\cite{10.1007/s11263-018-1112-4}, semantic segmentation~\cite{wei2016stc}, visual question answering (VQA)~\cite{sachan-xing-2016-easy}, and machine translation~\cite{platanios2019competence}. Although many curriculum learning approaches have been proposed for various tasks, to our knowledge, there are only a few studies that apply curriculum learning to video description~\cite{li2022adaptive}. 

Li et al.~\cite{li2022adaptive} leveraged curriculum learning to train a video description model based on the level of caption complexity and video diversity. 
Morerio et al.~\cite{morerio2017curriculum} proposed a curriculum learning approach to train an image classification model by dynamically increasing the dropout rate preventing the model from overfitting to specific samples.
Choi et al.~\cite{choi2019pseudo} apply curriculum learning in the unsupervised visual domain adaptation task. To improve image semantic segmentation, Wei et al.~\cite{wei2016stc} order the training data by the number of objects in images. Soviany et al.~\cite{soviany2020image} propose curriculum learning strategies for training generative adversarial networks (GANs) based on ranking the training images by their difficulty scores. Curriculum learning has been explored in VQA tasks to improve the model performance in answering complex questions about images~\cite{sachan-xing-2016-easy}. By starting with simpler questions that focus on basic visual attributes and gradually increasing the complexity of questions, models can learn to reason and answer more challenging queries.

Platanios et al.~\cite{platanios2019competence} order the training data from easy to hard based on sentence length and word rarity to improve machine translation. Kocmi et al.~\cite{kocmi2017curriculum} order the data according to linguistic phenomena, resulting in improved translation quality and convergence rate.

\section{Approach}

In this section, we present the details of our video description model, CLearViD. We discuss the underlying architecture of CLearViD and elaborate on the training process based on curriculum learning.

\subsection{Base Model Architecture}
We leveraged VLTinT~\cite{yamazaki2023vltint}, a transformer-based vision-language model that generates a paragraph description of untrimmed videos with annotated timestamped event segments. The model architecture consists of two main components: an encoder and a decoder. The encoder uses three modalities: (1) visual features of the scene extracted by a 3D-CNN model~\cite{ji20123d} pre-trained on Kinetcs-400~\cite{kay2017kinetics}, (2) visual features of the humans in the scene extracted by Faster-RCNN \cite{ren2015faster} pre-trained on the COCO dataset \cite{lin2014microsoft}, and (3)  linguistic features (i.e., a list of vocabulary) associated with the scene extracted by CLIP~\cite{pmlr-v139-radford21a} to capture both visual and non-visual elements. A fusion module models the interaction of these modalities and combines them into a unified representation. On the other hand, the decoder generates descriptions while maintaining semantic coherency between the paragraph sentences considering the video's intra-event and inter-event contents.

\subsection{Curriculum Learning by Noise}
We incorporated a curriculum learning mechanism into our model by gradually applying a Gaussian noise ($z$) to the video data during the training process. Gaussian noise follows a Gaussian distribution and its magnitude is directly proportional to the standard deviation ($\sigma$) of the distribution. Initially, $\sigma$ is set to zero at the start of training. Following the schedule shown in Figure \ref{fig:noise}, $\sigma$ is increased linearly during the early epochs. Once the $25^{th}$ epoch is reached, $\sigma$ remains fixed at its maximum value of $0.3$ for the remainder of the training process. The equation governing the dynamic $\sigma$ is as follows:

\begin{equation}
    \sigma = \mathrm{min}(\sigma_{max}, \sigma_{max}\times \frac{E}{E_{max}}),
\end{equation}

where $\sigma_{max}$ is the maximum standard deviation, $E$ is the current epoch, and $E_{max}$ is the epoch that the $\sigma$ reaches its maximum value. Curriculum learning by noise allows the model to learn complex features in the video data by first being exposed to simpler, less noisy versions of the data and gradually increasing the complexity. By progressively increasing the difficulty of the training examples, this method can lead to better convergence and increased resistance to overfitting.

\subsection{Curriculum Learning by Dropout}
\begin{figure}
\centering
\begin{subfigure}{0.45\textwidth}
     \includegraphics[width=\textwidth]{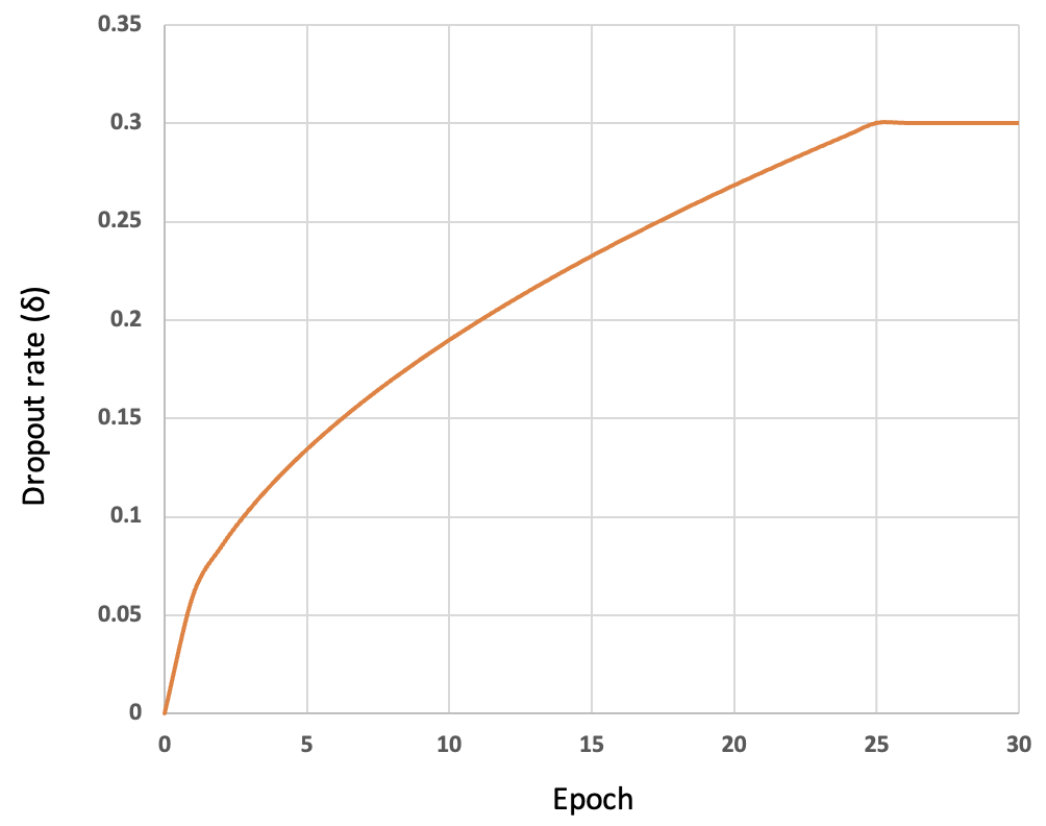}
     \caption{Curriculum Learning by Dropout}
     \label{fig:dropout}
\end{subfigure}
\hfill
\begin{subfigure}{0.45\textwidth}
    \includegraphics[width=\textwidth]{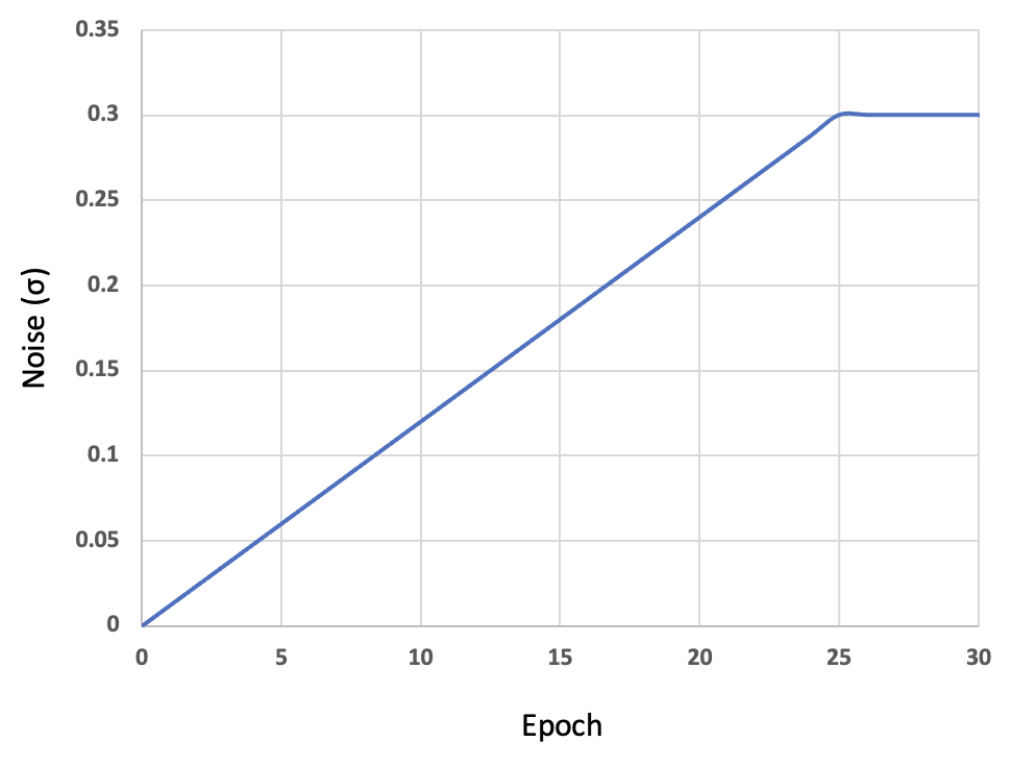}
    \caption{Curriculum Learning by Noise}
    \label{fig:noise}
\end{subfigure}
    \caption{Graphs of \textbf{(a)} the dropout rate, ($\delta$) in curriculum learning by dropout and \textbf{(b)} the standard deviation ($\sigma$) of the Gaussian noise in curriculum learning by noise during the training process.}
    \label{fig:cl-graph}
\end{figure}

We also leveraged another curriculum learning method by dynamically increasing the dropout rate ($\delta$) during training~\cite{morerio2017curriculum}. Initially, $\delta$ is set to zero, allowing the model to use all of its neurons. Following the schedule shown in Figure \ref{fig:dropout}, $\delta$ is increased during the early epochs. As training progresses and $\delta$ increases, the model has to learn the task with fewer neurons. By gradually reducing the capacity of the network through dropout, the model is encouraged to learn more robust and generalizable features. Once the $25^{th}$ epoch is reached, $\delta$ remains fixed at its maximum value of $0.3$ for the remainder of the training process. The equation governing the dynamic $\delta$ is as follows:

\begin{equation}
    \delta = \mathrm{min}(\delta_{max}, \delta_{max}\times \sqrt{\frac{E}{E_{max}}}),
\end{equation}

where $\delta_{max}$ is the maximum dropout rate, $E$ is the current epoch, and $E_{max}$ is the epoch at which the dropout rate reaches its maximum value.

\subsection{Mish Activation Function}
Mish \cite{misra2019mish} is a non-linear activation function that has been shown to improve performance in various machine learning tasks and mitigate issues associated with other activation functions, such as the vanishing gradient and the dead neuron problems. The Mish activation function has several desirable properties. For instance, it is continuously differentiable, which enables efficient gradient-based optimization during training. Moreover, it is non-monotonic, allowing for more complex representations and learning diverse patterns in data. Therefore, we replaced all instances of ReLU and GELU in the VLTinT model with Mish. The objective was to investigate whether Mish can improve the performance of the model. The Mish activation function is defined as follows:

\begin{equation}
\mathrm{Mish}(x) = x \times \tanh(\ln(1+e^x)),
\end{equation}

where $x$ is the input to the activation function.

\section{Experiments and Results}

\subsection{Datasets}
We evaluated CLearViD on two popular datasets, i.e., ActivityNet
Captions~\cite{krishna2017dense} and YouCook2~\cite{zhou2018towards}. ActivityNet Captions consists of $19,994$ videos, with $10,024$ videos in the training set, $4,926$ videos in the validation set with two reference descriptions (referred to as \textit{ae-val}), and $5,044$ videos in the test set (referred to as \textit{ae-test})~\cite{krishna2017dense}. The dataset provides timestamps for event segments in the videos with start and end times. Following previous works \cite{gella2018dataset, zhou2018end, park2019adversarial, kanani2021global}, we used the second reference description of the validation set for training CLearViD and the first reference description for evaluation. ActivityNet Captions is one of the largest and most diverse datasets available for video description. YouCook2 also consists of $1,790$ videos, with $457$ videos in the validation set and $1,333$ videos in the training set. Each video in the dataset contains $7.7$ event segments on average. Unlike ActivityNet Captions, YouCook2 has only one reference description in the validation set for evaluation. 

\subsection{Implementation Details} \label{Implementation_details}

CLearViD follows the specific configurations used in the VLTinT model \cite{yamazaki2023vltint}: a hidden size of 768, 3 transformer layers, and 12 attention heads. We used Adam optimizer to train the model and start with an initial learning rate of $\alpha=1e-4$, $\beta1 = 0.9$, $\beta2 = 0.99$, and $L2$ weight decay of $0.01$. We applied learning rate warm-up over the first five epochs. To mitigate overfitting, we used label smoothing with $\epsilon = 0.1$ and $\lambda = 0.1$. For curriculum learning by dropout, we set the maximum dropout rate, $\delta_{max}$, to $0.25$. For curriculum learning by noise, we set the maximum standard deviation, $\sigma_{max}$, to $0.3$. We used a fixed random seed value of $2019$ to ensure the reliability and validity of our experimental results. The training process was performed on a single NVIDIA A100 (80GB) GPU.

\begin{table}
    \caption{Performance comparison of CLearViD with other SOTA models on ActivityNet Captions \textit{ae-val}. `-' denotes the paper did not report the score for that particular metric.}
    \begin{tabular*}{\linewidth}{@{\extracolsep{\fill}} l|cccc|cc }
        \toprule
        Model           & METEOR↑  & ROGUE\_L↑  & CIDEr↑ & Bleu@4↑ & Div2↑ & RE-4↓ \\ \midrule
        Vanilla Trans. \cite{zhou2018end}  & 15.64 & 28.90 & 22.16 & 9.75  & 77.40 & 7.79  \\
        Adv-inf \cite{park2019adversarial}        & 16.60 & -     & 20.97 & 10.04 & -     & 5.76  \\
        GVD \cite{zhou2019grounded}            & 15.71 & -     & 21.95 & 11.04 & -     & 8.76  \\
        Trans.-XL  \cite{dai2019transformer} & 15.09 & 30.18 & 21.67 & 10.39 & 75.96 & 8.54  \\
        Trans.-XLRG \cite{lei2020mart}    & 14.77 & -     & 20.40 & 10.17 & -     & 8.85  \\
        MART  \cite{lei2020mart}           & 15.68 & 30.32 & 23.42 & 10.33 & 75.71 & 5.18  \\
        PDVC \cite{wang2021end}           & 15.93 & -     & 23.42 & 11.80 & -     & -     \\
        %VLCAP \cite{yamazaki2022vlcap}          & 17.78 & 36.37 & 32.58 & 14    & 78.01 & \textbf{4.42}  \\
        VLTinT  \cite{yamazaki2023vltint}        & 18.16 & 36.86 & 33.07 & 14.93 & 77.72 & 4.87  \\
        \bottomrule
        \textbf{CLearViD} (Ours)  & \textbf{18.18} & \textbf{37.37} & \textbf{34.41} & \textbf{15.22} & \textbf{78.50} & \textbf{4.53}  \\
        \bottomrule
        \end{tabular*}
    \label{tab:anet-val}
\end{table}

\begin{table}
    \caption{Performance comparison of CLearViD with other SOTA models on ActivityNet Captions \textit{ae-test}. `-' denotes the paper did not report the score for that particular metric.}
    \begin{tabular*}{\linewidth}{@{\extracolsep{\fill}} l|cccc|cc }
        \toprule
        Model      & METEOR↑   & ROGUE\_L↑ & CIDEr↑    & Bleu@4↑   & Div2↑ & RE-4↓  
        \\ \midrule
        Vanilla Trans. \cite{zhou2018end}     & 15.54 & 28.98 & 21.33 & 9.31  & 77.29 & 7.45   \\
        Trans.-XL \cite{dai2019transformer}          & 14.91 & 30.25 & 21.71 & 10.25 & 76.17 & 8.79   \\
        Trans.-XLRG \cite{lei2020mart}        & 14.58 & -     & 20.34 & 10.07 & -     & 9.37   \\
        MART  \cite{lei2020mart}              & 15.57 & 30.85 & 22.16 & 9.78  & 75.69 & 5.44   \\
        MART$^{COOT}$ \cite{ging2020coot}        & 15.99 & -     & 28.19 & 10.85 & -     & 6.64   \\
        Memory Trans. \cite{song2021towards}      & 15.64 & -     & 26.55 & 11.74 & \textbf{83.95} & \textbf{2.75}   \\
        VLTinT \cite{yamazaki2023vltint} & 17.97 & 36.56 & 31.13 & 14.50 & 77.72 & 4.75 \\
        \bottomrule
        %CL by dropout + Mish
        \textbf{CLearViD} (Ours) & \textbf{18.03} & \textbf{36.67} & \textbf{32.95} & \textbf{14.62} & 78.82 & 4.01   \\
        \bottomrule
        \end{tabular*}
    \label{tab:anet-test}
\end{table}

\begin{table}[t!]
    \caption{Performance comparison of CLearViD with other SOTA models on YouCook2. `-' denotes the paper did not report the score for that particular metric. $^\dagger$ represents results generated by our team.} 
    \begin{tabular*}{\linewidth}{@{\extracolsep{\fill}} l|cccc|cc }
        \toprule
        Model  & METEOR↑ & ROGUE\_L↑  & CIDEr↑  & Bleu@4↑ & Div2↑ & RE-4↓  \\ 
        \midrule
        Vanilla Trans. \cite{zhou2018end} & 11.55 & -     & 38.00 & 4.38 & - & -     \\
        MART  \cite{lei2020mart}          & 15.90 & -     & 35.74 & 8.00 & - & 4.39  \\
        MART$^{COOT}$ \cite{ging2020coot}    & \textbf{18.17} & -     & 46.06 & 9.44 & - & 6.30  \\
        VLTinT \cite{yamazaki2023vltint}         & 17.94 & 34.55 & 48.70 & 9.40 & 67.36$^\dagger$ & 4.29  \\ 
        \bottomrule
        \textbf{CLearViD} (Ours) & 17.89 & \textbf{35.24} & 4\textbf{8.89} & \textbf{9.92} & \textbf{70.15} & \textbf{4.27} \\
        \bottomrule
        \end{tabular*}
    \label{tab:yc2}
\end{table}

\subsection{Quantitative Analysis}
We conducted a performance comparison between CLearViD and several SOTA models in the literature for the video description task. Our evaluation includes commonly used metrics such as METEOR~\cite{denkowski2014meteor}, ROGUE\_L~\cite{lin2004rouge}, BLEU@4~\cite{papineni2002bleu}, and CIDEr-D~\cite{vedantam2015cider}. To account for the diversity of content or repetition of phrases and sentence structures, we also present Div-2 scores~\cite{shetty2017speaking}, which measure the ratio of unique $n$-grams ($n=1, 2$) to the total number of words, and RE-4~\cite{xiong2018move}, which captures the degree of $n$-gram repetition ($n=4$) in a description. For each metric, we compute the score for each video and report the average score across all videos in the dataset. We present the results on ActivityNet Captions \textit{ae-val} (Table \ref{tab:anet-val}), ActivityNet Captions \textit{ae-test} (Table \ref{tab:anet-test}), and YouCook2 (Table \ref{tab:yc2}). CLearViD trained with a mixed curriculum of noise and dropout and the Mish activation function outperformed the SOTA in all accuracy and diversity metrics on both datasets.

\begin{figure}[t!]
\centering
\begin{subfigure}[t]{0.32\textwidth}
     \includegraphics[width=\textwidth]{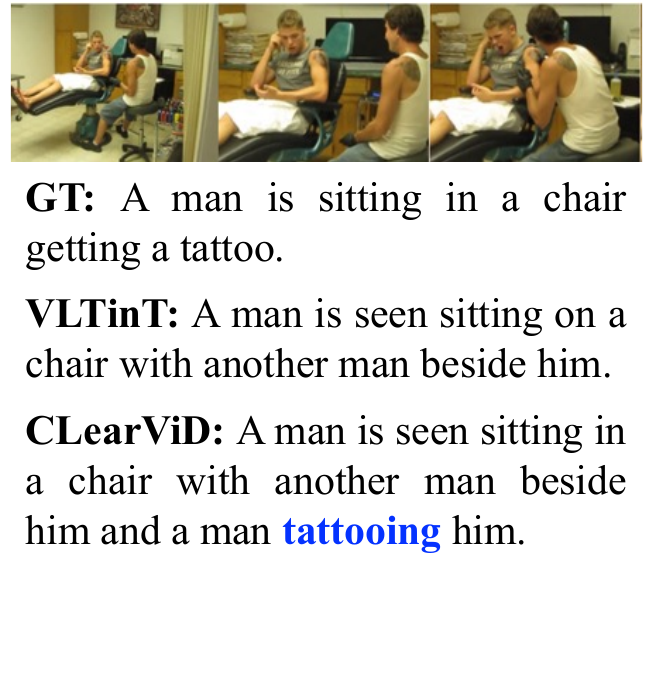}
     \caption{ActivityNet Captions \textit{ae-val}}
     \label{fig:visual_result_a}
\end{subfigure}
\vspace{0.3cm}
\begin{subfigure}[t]{0.32\textwidth}
     \includegraphics[width=\textwidth]{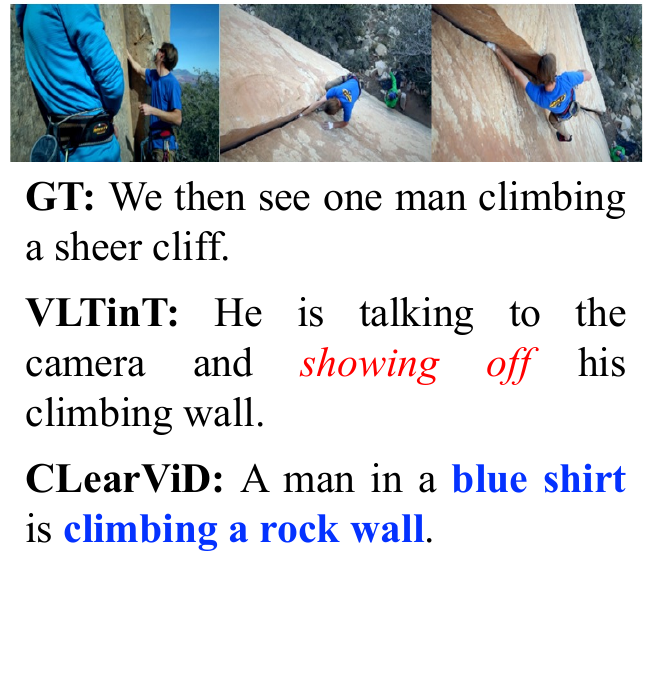}
     \caption{ActivityNet Captions \textit{ae-val}}
     \label{fig:visual_result_b}
\end{subfigure}
\begin{subfigure}[t]{0.32\textwidth}
    \includegraphics[width=\textwidth]{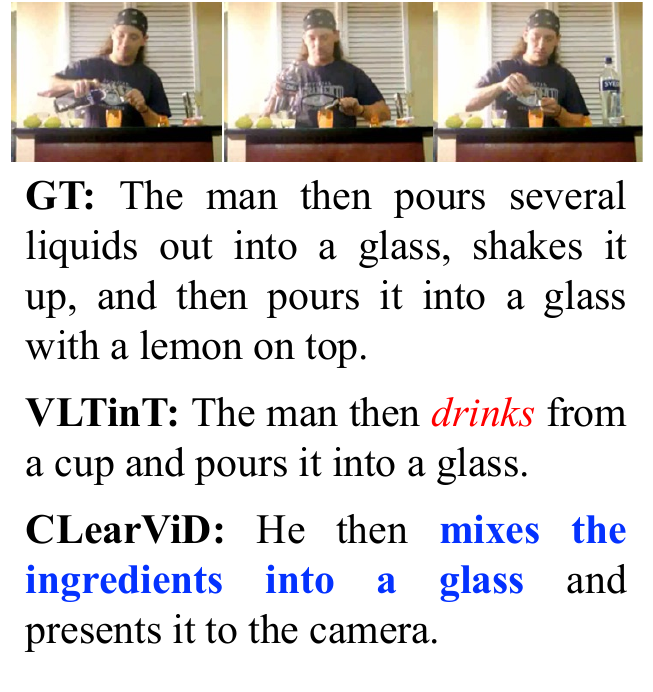}
    \caption{ActivityNet Captions \textit{ae-test}}
    \label{fig:visual_result_c}
\end{subfigure}
\begin{subfigure}{0.32\textwidth}
     \includegraphics[width=\textwidth]{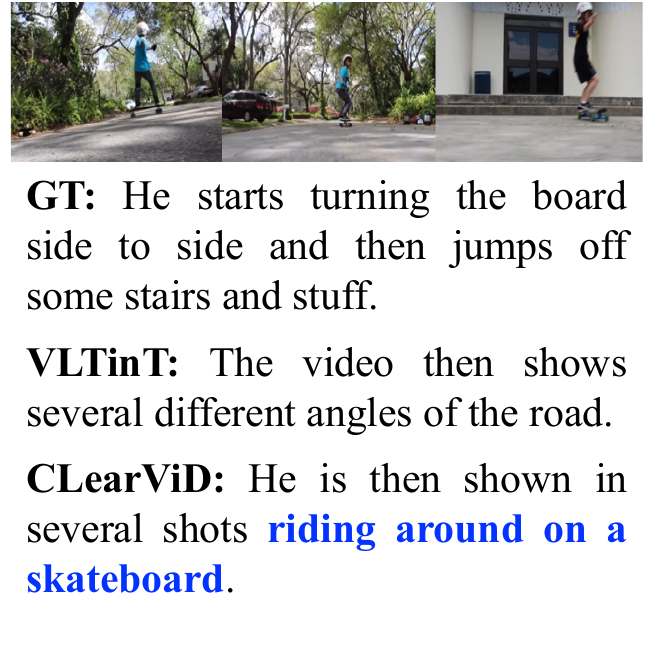}
     \caption{ActivityNet Captions \textit{ae-test}}
     \label{fig:visual_result_d}
\end{subfigure}
\vspace{0.3cm}
\begin{subfigure}{0.32\textwidth}
     \includegraphics[width=\textwidth]{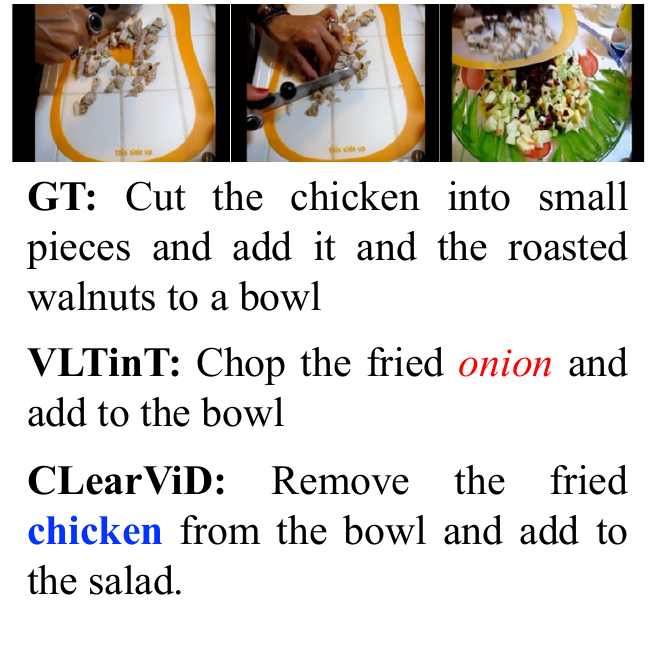}
     \caption{YouCook2}
     \label{fig:visual_result_e}
\end{subfigure}
\begin{subfigure}{0.32\textwidth}
    \includegraphics[width=\textwidth]{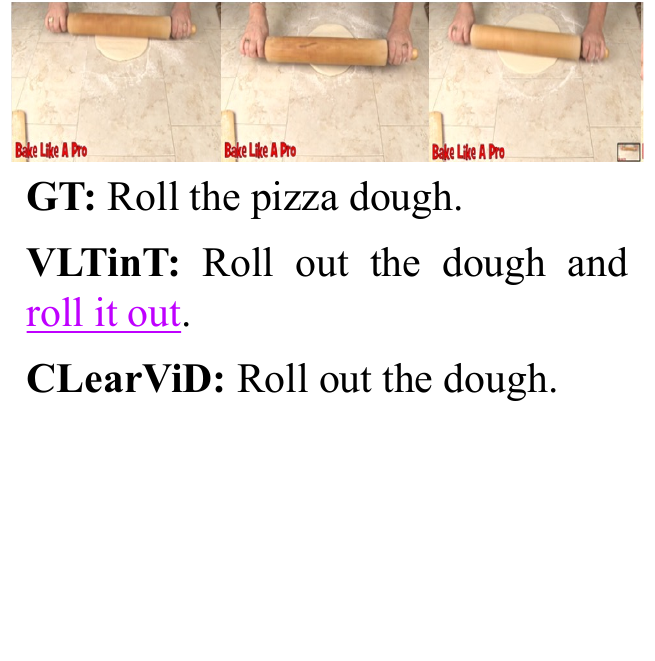}
    \caption{YouCook2}
    \label{fig:visual_result_f}
\end{subfigure}
\begin{subfigure}{0.96\textwidth}
    \includegraphics[width=\textwidth]{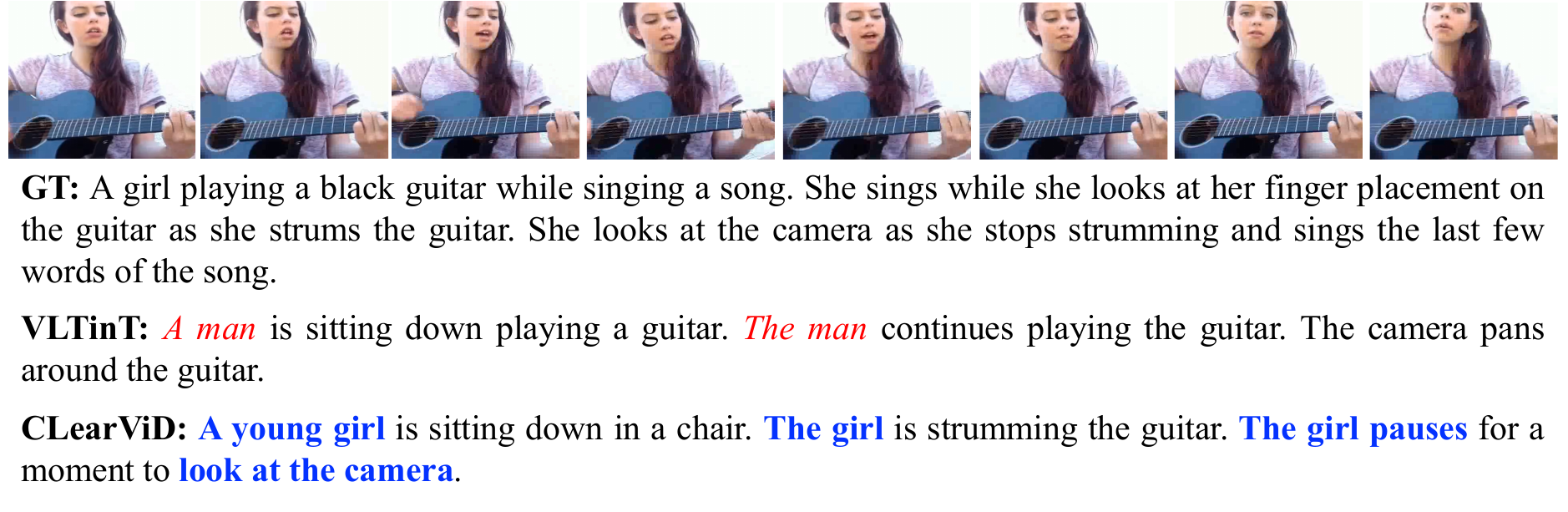}
    \caption{ActivityNet Captions \textit{ae-val}}
    \label{fig:visual_result_g}
\end{subfigure}
    \caption{Qualitative comparison of the ground truth (GT) descriptions with those generated by the VLTinT baseline model and our proposed model, CLearViD. The examples are from the ActivityNet Captions \textit{ae-val}, ActivityNet Captions \textit{ae-test}, and YouCook2 datasets. \textit{\textcolor{red}{Red italics}} indicates the description errors, \textcolor{electricpurple}{\underline{purple underlining}} indicates repetitive patterns, and \textbf{\textcolor{blue}{blue bold}} indicates distinct expressions.}
    \label{fig:visual-result}
\vspace{-0.2cm}
\end{figure}

The results on ActivityNet Captions \textit{ae-val} demonstrate the effectiveness of our video description model, and we observed improvements in all six metrics compared to the best SOTA scores, namely, METEOR (+0.11\%), CIDEr (+4.05\%), ROGUE\_L (+1.38\%), BLEU@4 (+1.94\%), Div2 (+1.00\%), and RE-4 (-6.98\%). We also observed similar improvements on ActivityNet Captions \textit{ae-test}: METEOR (+0.33\%), CIDEr (+5.85\%), ROGUE\_L (+0.30\%), and BLEU@4 (+0.83\%). On the YouCook2 dataset, CLearViD also demonstrates a significant superiority over existing SOTA models in terms of both accuracy and diversity metrics: CIDEr (+0.39\%), ROGUE\_L (+2.00\%), Bleu@4 (+5.53\%), Div2 (+4.14\%), and RE-4 (-0.47\%).

\subsection{Qualitative Analysis}
While evaluating the video descriptions generated by CLearViD, we observed that they were more detailed and robust compared to those produced by the baseline models. The descriptions expressed the essential parts of the video concisely and clearly while providing additional context that improved the viewer's comprehension of the video content. Figure~\ref{fig:visual-result} illustrates the visual results of CLearViD compared to the baseline model, VLTinT, and the ground truth (GT). It demonstrates that CLearViD generates more distinct and accurate descriptions while reducing redundancy. For instance, in Figure~\ref{fig:visual_result_a}, CLearViD correctly identifies that a man is getting a tattoo, while the baseline model misses it. In Figure~\ref{fig:visual_result_b}, CLearViD not only identifies that a man is climbing a rock wall but also recognizes that he is wearing a blue shirt, providing more detail. In Figure~\ref{fig:visual_result_c}, while the baseline model makes an error, CLearViD correctly recognizes that the man is not drinking. In Figure~\ref{fig:visual_result_d}, CLearViD accurately identifies the type of activity taking place in the video. In Figure~\ref{fig:visual_result_e}, CLearViD identifies the correct ingredient as chicken, rather than onion. In Figure~\ref{fig:visual_result_f}, CLearViD effectively reduces content repetition, and in Figure~\ref{fig:visual_result_g}, CLearViD identifies the correct gender in the video and captures the nuanced movement when the girl pauses and looks at the camera.

\begin{table}[t!]
  \caption{Results of ablation studies on the effectiveness of each approach on ActivityNet Captions \textit{ae-val}} %We compare our results with the SoTA model: VLTinT.}
  \centering
  \begin{tabular}{ccc|cccc|cc}
    \toprule
    Noise & Dropout & Mish  & METEOR↑  & ROGUE\_L↑  & CIDEr↑  & Bleu@4↑ & Div2↑ & RE-4↓ \\
    \midrule
     ✗ & ✗ & ✗  & 18.16 & 36.86 & 33.07 & 14.93 & 77.72 & 4.87  \\
     ✓ & ✗ & ✗  & 18.19 & 37.28 & 34.03 & 15.12 & 78.38 & 4.75  \\
     ✗ & ✓ & ✗  & 18.15 & 37.32 & 33.60 & 15.15 & 77.81 & 5.06  \\
     ✗ & ✗ & ✓  & 18.11 & 37.14 & 33.57 & 15.05 & 77.84 & 5.25  \\
     ✓ & ✗ & ✓  & 18.17 & 37.25 & 34.00 & 15.08 & 77.83 & 5.24  \\
     ✗ & ✓ & ✓  & 18.14 & 37.21 & 34.12 & 15.11 & 78.62 & 4.72  \\
     ✓ & ✓ & ✗  & \textbf{18.23} & 37.24 & \textbf{34.50} & 15.20 & \textbf{78.63} & 4.66  \\
     ✓ & ✓ & ✓  & 18.18 & \textbf{37.37} & 34.41 & \textbf{15.22} & 78.50 & \textbf{4.53}  \\
    \bottomrule
  \end{tabular}
  \label{tab:ab-anet-val}
  \vspace{-0.01cm}
\end{table}

\begin{table}[t!]
  \caption{Results of ablation studies on the effectiveness of each approach on ActivityNet Captions \textit{ae-test}.}
  \centering
  \begin{tabular}{ccc|cccc|cc}
    \toprule
    Noise & Dropout & Mish  & METEOR↑  & ROGUE\_L↑  & CIDEr↑  & Bleu@4↑ & Div2↑ & RE-4↓ \\
    \midrule
    ✗ & ✗ & ✗  & 17.97 & 36.56 & 31.13 & 14.50 & 77.72 & 4.75  \\
    ✓ & ✗ & ✗  & 17.96 & 36.76 & 30.37 & \textbf{14.71} & 77.89 & 5.59  \\
    ✗ & ✓ & ✗  & 17.91 & 36.58 & 31.26 & 14.65 & 77.73 & 5.23  \\
    ✗ & ✗ & ✓  & 17.84 & 36.60 & 30.77 & 14.64 & 76.21 & 6.10  \\
    ✓ & ✗ & ✓  & 17.95 & 36.88 & 31.37 & 14.51 & 76.32 & 6.07  \\
    ✗ & ✓ & ✓  & 17.98 & 36.96 & 31.62 & 14.58 & 77.81 & 5.26  \\
     ✓ & ✓ & ✗  & 17.84 & \textbf{36.98} & 32.18 & 14.70 & 78.09 & 4.51  \\
     ✓ & ✓ & ✓  & \textbf{18.03} & 36.67 & \textbf{32.95} & 14.62 & \textbf{78.82} & \textbf{4.01}  \\
    \bottomrule
  \end{tabular}
  \label{tab:ab-anet-test}
    \vspace{-0.01cm}
\end{table}

\begin{table}[t!]
  \caption{Results of ablation studies on the effectiveness of each approach on YouCook2. $^\dagger$ represents results generated by our team.}
  \centering
  \begin{tabular}{ccc|cccc|cc}
    \toprule
    Noise & Dropout & Mish  & METEOR↑  & ROGUE\_L↑  & CIDEr↑  & Bleu@4↑ & Div2↑ & RE-4↓ \\
    \midrule
    ✗ & ✗ & ✗   & 17.94 & 34.55 & 48.70 & 9.40 & 67.36$^\dagger$ & 4.29 \\
    ✓ & ✗ & ✗   & \textbf{17.99} & 34.73 & 49.19 & 9.41 & 69.89 & 4.13 \\
    ✗ & ✓ & ✗   & 17.92 & 34.74 & 48.80 & 9.60 & \textbf{70.33} & 4.20 \\
    ✗ & ✗ & ✓   & 17.73 & 34.57 & 46.02 & 9.52 & 68.95 & 5.32 \\
    ✓ & ✗ & ✓   & 17.67 & 34.91 & \textbf{49.95} & 9.59 & 69.64 & 4.18 \\
    ✗ & ✓ & ✓   & 17.82 & 35.10 & 49.44 & 9.74 & 69.79 & 4.60 \\
     ✓ & ✓ & ✗  & 17.92 & 34.88 & 49.43 & 9.90 & 69.82 & \textbf{3.98}  \\
     ✓ & ✓ & ✓  & 17.89 & \textbf{35.24} & 48.89 & \textbf{9.92} & 70.15 & 4.27  \\
    \bottomrule
  \end{tabular}
  \label{tab:ab-yc2}
\end{table}

\subsection{Ablation Studies}
We conducted a series of ablation studies to evaluate the effectiveness of our proposed model. We ran experiments with different settings to assess various components of the model's performance across the datasets, i.e., ActivityNet Captions \textit{ae-val} (Table \ref{tab:ab-anet-val}), ActivityNet Captions \textit{ae-test} (Table \ref{tab:ab-anet-test}), and YouCook2 (Table \ref{tab:ab-yc2}). The first row in the tables shows the results for CLearViD's base model architecture (i.e., VLTinT), which does not employ curriculum learning or the Mish activation function. The next three rows show the performance and contribution of each individual approach (i.e., curriculum learning by noise, curriculum learning by dropout, and the Mish function) in CLearViD, whereas the last four rows show the performance of different combinations. The best performance overall is obtained by combining all three methods.

\paragraph{Effectiveness of Curriculum Learning By Noise.} Curriculum learning by noise demonstrates strong performance on ActivityNet Captions \textit{ae-val} and YouCook2, resulting in improved scores across all six accuracy and diversity metrics compared to the base model. However, its performance on ActivityNet Captions \textit{ae-test} varies. When combined with curriculum learning by dropout, the method further enhances the scores on the majority of metrics on both datasets. As demonstrated in the Appendix, the gradual and increasing application of Gaussian noise to the video data during training, as opposed to applying a fixed noise, results in improved convergence and overall performance on evaluation metrics.

\paragraph{Effectiveness of Curriculum Learning by Dropout.} Curriculum learning by dropout improves the model performance on both datasets. When combined with curriculum learning by noise, the method further enhances the scores on the majority of metrics. However, when combined with Mish, its performance varies.

\paragraph{Effectiveness of the Mish Activation Function.} We replaced all instances of ReLU and GELU activation functions in the base model, VLTinT, with Mish. The conducted ablation studies show that Mish alone does not enhance performance compared to ReLU and GELU. However, when combined with both curriculum learning by noise and dropout, Mish outperforms ReLU and GELU on accuracy and diversity metrics.

\section{Limitations}
Video description models are still in their infancy, and these models are notoriously data hungry. The datasets we need to train our model must be significantly larger and more diverse than what was used in this work to match the variety of vocabulary, scenes, and contexts in videos. These models are also typically trained on datasets that contain human-generated descriptions. Annotators might have implicit biases that influence their choice of words or the way they describe certain scenes or actions in the videos. These biases can propagate into the training data and subsequently influence the output of the models. Moreover, if the training data contains an overrepresentation of certain races or gender roles, the models might disproportionately describe or focus on those aspects in the generated descriptions. Therefore, curating diverse and balanced training datasets that represent various demographics, cultures, and perspectives is crucial for mitigating biases. Other methods, such as debiasing, re-ranking, adversarial training, or fairness-aware evaluation metrics can also be employed to mitigate biases in the model outputs.

\section{Conclusion} \label{Conclusion}
We introduced CLearViD, a novel approach for automatic video description generation. The proposed model leverages curriculum learning and the Mish activation function to accomplish this task. 
Our comprehensive experiments and ablation studies validated the effectiveness of the proposed model, and the results showed that CLearViD outperforms existing SOTA models on both accuracy and diversity metrics.
In our future work, we plan to explore distillation techniques~\cite{gou2021knowledge} to enhance computational efficiency and optimize the model for portability and real-time processing.
In addition, we will investigate sample-efficient (i.e., learning with less data) models for video description by leveraging self-supervised learning methods~\cite{doersch2017multi, wang2015unsupervised}. Lastly, we emphasize the importance of incorporating ethical guidelines and practices into the development and deployment of video description models to promote their inclusive use across various domains.

\bibliography{references}

\newpage

\appendix

\section{Appendix}

\subsection{Fixed vs.\ Scheduled Noise} \label{fixed-noise}

We conducted a set of experiments to train the base VLTinT model~\cite{yamazaki2023vltint} using a fixed $\sigma$ for the Gaussian noise. Instead of dynamically adjusting the $\sigma$ according to a curriculum, we used a fixed $\sigma$, ranging from $0.1$ to $0.5$, throughout the training process. We compared this approach with curriculum learning by noise, which employs a scheduled $\sigma$ (Figure~\ref{fig:noise}), to assess the significance of curriculum-based learning in model performance. The results on the ActivityNet Captions and YouCook2 datasets (Tables \ref{tab:fix-noise-anet-val}, \ref{tab:fix-noise-anet-test}, \ref{tab:fix-noise-yc2}) show that fixed noise impedes the learning process, and curriculum learning by noise effectively mitigates the negative effects of fixed noise while improving accuracy and diversity of descriptions. 

\begin{table*}[h]
    \caption{ Performance comparison of a fixed vs.\ scheduled $\sigma$ of Gaussian noise on ActivityNet Captions \textit{ae-val}. CL: curriculum learning.}
    \begin{tabular*}{\linewidth}
    {@{\extracolsep{\fill}} c|c|cccc|cc }
        \toprule
        Approach & $\sigma$  & METEOR↑ & ROGUE\_L↑  & CIDEr↑  & Bleu@4↑ & Div2↑ & RE-4↓  \\ 
        \midrule
        % VLTinT & N/A & 18.16 & 36.86 & 33.07 & 14.93 & 77.72 & 4.87  \\ 
        CL by Noise & Scheduled & \textbf{18.19} & \textbf{37.28} & \textbf{34.03} & 15.12 & \textbf{78.38} & \textbf{4.75}  \\
        \midrule
                     & 0.1 & 18.06 & 37.22 & 32.08 & \textbf{15.14} & 65.92 & 08.22 \\
                     & 0.2 & 17.41 & 36.48 & 27.49 & 13.95 & 64.81 & 10.65 \\
        Fixed Noise & 0.3 & 17.38 & 36.52 & 27.28 & 14.03 & 64.77 & 10.70 \\
                     & 0.4 & 17.35 & 36.48 & 27.53 & 13.99 & 64.76 & 10.71 \\
          & 0.5 & 17.33 & 36.55 & 27.20 & 14.04 & 64.69 & 10.61 \\
                     % & 0.6 & 17.34 & 36.58 & 27.03 & 14.02 & 64.73 & 10.55 \\
                     % & 0.7 & 17.30 & 36.52 & 27.24 & 13.99 & 64.83 & 10.47 \\
                     % & 0.8 & 17.33 & 36.46 & 26.91 & 13.87 & 64.77 & 10.83 \\
                     % & 0.9 & 17.27 & 36.51 & 26.45 & 13.89 & 64.61 & 10.65 \\
                     % & 1.0 & 17.29 & 36.44 & 26.18 & 13.83 & 64.72 & 10.76 \\
        \bottomrule
        \end{tabular*}
    \label{tab:fix-noise-anet-val}
\end{table*}

\begin{table*}[h]
    \caption{ Performance comparison of a fixed vs.\ scheduled $\sigma$ of Gaussian noise on ActivityNet Captions \textit{ae-test}. CL: curriculum learning.}
   
    \begin{tabular*}{\linewidth}
    {@{\extracolsep{\fill}} c|c|cccc|cc }
        \toprule
        Approach & $\sigma$  & METEOR↑ & ROGUE\_L↑  & CIDEr↑  & Bleu@4↑ & Div2↑ & RE-4↓  \\ 
        \midrule
        % VLTinT & N/A & 17.97 & 36.56 & 31.13 & 14.50 & 77.72 & 4.75 \\
        CL by Noise & Scheduled & \textbf{18.03} & \textbf{36.67} & \textbf{32.95} & \textbf{14.62} & \textbf{78.82} & \textbf{4.01}   \\
        \midrule
                    & 0.1 & 16.89 & 36.54 & 29.55 & 14.04 & 65.14 & 7.84  \\
                    & 0.2 & 16.11 & 35.68 & 24.61 & 12.41 & 63.79 & 10.37 \\
        Fixed Noise  & 0.3 & 16.22 & 35.89 & 24.34 & 12.57 & 63.90 & 10.47 \\
                    & 0.4 & 15.85 & 35.87 & 24.99 & 12.71 & 63.88 & 10.56 \\
         & 0.5 & 15.74 & 35.92 & 24.58 & 12.79 & 63.98 & 10.64 \\
                    % & 0.6 & 16.05 & 35.64 & 24.44 & 12.55 & 64.22 & 10.20 \\
                    % & 0.7 & 15.88 & 35.74 & 24.71 & 12.63 & 64.21 & 10.12 \\
                    % & 0.8 & 15.93 & 35.71 & 24.30 & 12.52 & 63.97 & 10.76 \\
                    % & 0.9 & 15.70 & 35.87 & 23.82 & 12.72 & 63.74 & 10.39 \\
                    % & 1.0 & 15.69 & 35.75 & 23.19 & 12.58 & 63.72 & 10.46 \\
        \bottomrule
        \end{tabular*}
    \label{tab:fix-noise-anet-test}
\end{table*}

\begin{table*}[h]
    \caption{Performance comparison of a fixed vs.\ scheduled $\sigma$ of Gaussian noise on the YouCook2 dataset. CL: curriculum learning.}
    \begin{tabular*}{\linewidth}
    {@{\extracolsep{\fill}} c|c|cccc|cc }
        \toprule
        Approach & $\sigma$  & METEOR↑ & ROGUE\_L↑  & CIDEr↑  & Bleu@4↑ & Div2↑ & RE-4↓  \\ 
        \midrule
        % VLTinT & N/A & 17.94 & 34.33 & 48.70 & 9.40  & 4.29  \\ 
        CL by Noise & Scheduled & \textbf{17.99} & \textbf{34.73} & \textbf{49.19} & \textbf{9.41} & \textbf{69.89}  & \textbf{4.13}  \\
        \midrule
                  & 0.1 & 17.23 & 34.62 & 43.86 & 9.23 & 65.89 & 6.27 \\
                  & 0.2 & 17.00 & 34.41 & 43.85 & 9.05 & 64.75 & 7.03 \\
        Fixed Noise  & 0.3 & 16.98 & 34.19 & 43.28 & 9.08 & 64.48 & 7.13 \\
                  & 0.4 & 16.99 & 34.04 & 41.17 & 9.03 & 65.04 & 7.30 \\
         & 0.5 & 17.18 & 34.08 & 44.00 & 9.36 & 66.44 & 6.29 \\

        \bottomrule
        \end{tabular*}
    \label{tab:fix-noise-yc2}
\end{table*}

\subsection{Assets We Used} \label{ap-assets}
Here is the list of assets we use in this work:

\begin{itemize}
    \item VLTinT: None, public GitHub repository 
    \item Densevid\_eval: MIT 
    \item CocoCaption: Creative Commons Attribution 4.0 International (CC BY 4.0) license 
    \item SlowFast: Apache License 2.0 
    \item Detectron2: Apache License 2.0 
    \item ActivityNet Captions: Creative Commons Attribution 4.0 International (CC BY 4.0) license 
    \item YouCook2: Creative Commons Attribution-NonCommercial-ShareAlike 3.0 (CC BY-NC-SA 3.0) license 
\end{itemize}

\subsection{Visual Results}

\begin{figure}[h]
    \centering
    \includegraphics[width=0.96\textwidth]{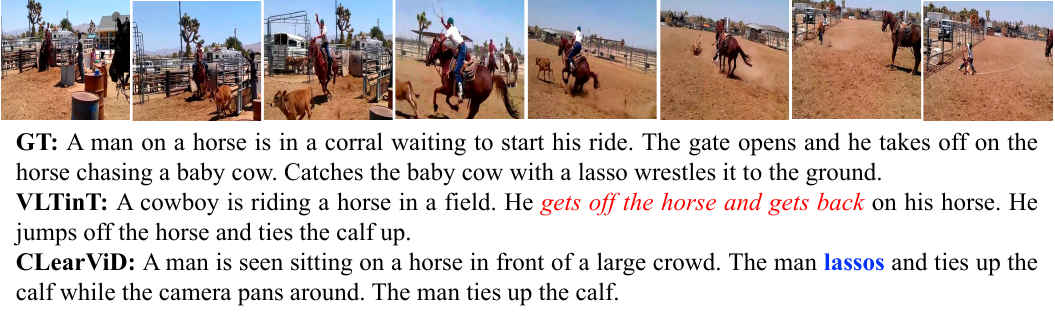}
    \includegraphics[width=0.96\textwidth]{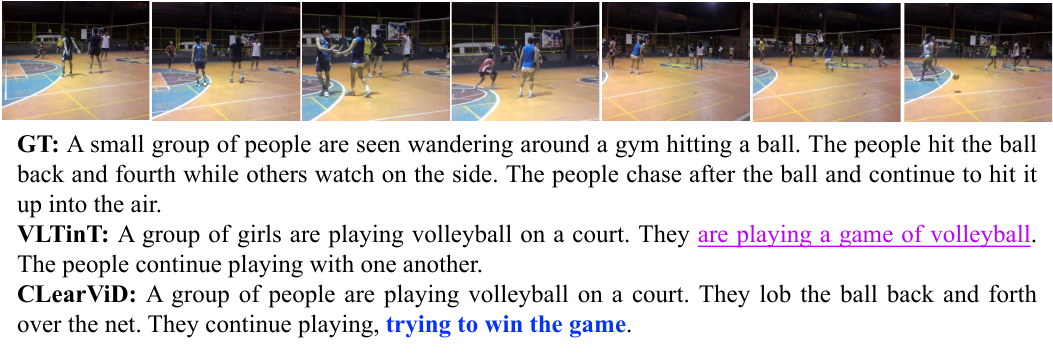}
    \includegraphics[width=0.96\textwidth]{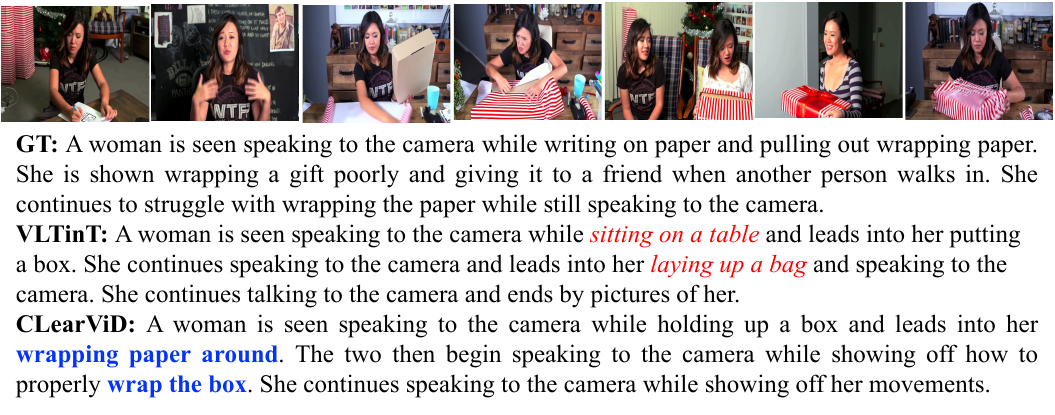}
    % \centering
    % \includegraphics[width=0.96\textwidth]{Appendix_VisualResults_3.pdf} % the beach result has too many errors in GT
    % \includegraphics[width=0.96\textwidth]{Appendix_VisualResults_4.pdf}
    \caption{Qualitative comparison of the ground truth (GT) descriptions with those generated by the VLTinT baseline model and our proposed model, CLearViD. The examples are from the  ActivityNet Captions \textit{ae-test} dataset. \textit{\textcolor{red}{Red italics}} indicates the description errors, \textcolor{electricpurple}{\underline{purple underlining}} indicates repetitive patterns, and \textbf{\textcolor{blue}{blue bold}} indicates distinct expressions.}
    % \vspace*{6in} % move figure to top of the page
    \label{fig:ap-visual-results}
\end{figure}

%\subsection{Visual Results} \label{ap-visual-results}
Figure~\ref{fig:ap-visual-results} illustrates supplementary visual results for qualitative comparison between the ground truth (GT) descriptions, those generated by the VLTinT baseline model, and our proposed model, CLearViD. CLearVid employs curriculum learning by noise, curriculum learning by dropout, and the Mish activation function. 

\end{document}